\documentclass[journal]{IEEEtran}
\usepackage{graphicx, latexsym}
\usepackage{subfigure,amsmath,amssymb,caption,float,color,cite,epsfig}
\usepackage{algorithmic}
\usepackage{hyperref,MnSymbol}

\newtheorem{theorem}{Theorem}

\newtheorem{corollary}{Corollary}
\newtheorem{lemma}{Lemma}

\def \1{{\bf 1}}

\ifCLASSINFOpdf
\else
\fi

\begin{document}
%
\title{Efficient Adaptive Compressive Sensing Using Sparse Hierarchical Learned Dictionaries}

\author{\IEEEauthorblockN{Akshay Soni and Jarvis Haupt
\thanks{This work was supported by DARPA/ONR under Award No. N66001-11-1-4090.}}\\
\IEEEauthorblockA{University of Minnesota, Twin Cities\\
Department of Electrical and Computer Engineering\\
Minneapolis, Minnesota USA 55455\\
e-mail: {\tt \{sonix022,jdhaupt\}@umn.edu}}

}


\maketitle

\begin{abstract}
Recent breakthrough results in compressed sensing (CS) have established that many high dimensional objects can be accurately recovered from a relatively small number of non-adaptive linear projection observations, provided that the objects possess a sparse representation in some basis.  Subsequent efforts have shown that the performance of CS can be improved by exploiting the structure in the location of the non-zero signal coefficients (structured sparsity) or using some form of online measurement focusing (adaptivity) in the sensing process. In this paper we examine a powerful hybrid of these two techniques.  First, we describe a simple adaptive sensing procedure and show that it is a provably effective method for acquiring sparse signals that exhibit structured sparsity characterized by tree-based coefficient dependencies.  Next, employing techniques from sparse hierarchical dictionary learning, we show that representations exhibiting the appropriate form of structured sparsity can be learned from collections of training data.  The combination of these techniques results in an effective and efficient adaptive compressive acquisition procedure.

\end{abstract}


%
\IEEEpeerreviewmaketitle

\section{Introduction}
Motivated in large part by breakthrough results in compressed sensing (CS), significant attention has been focused in recent years on the development and analysis of sampling and inference methods that make efficient use of measurement resources. The essential idea underlying many directions of research in this area is that signals of interest often possess a parsimonious representation in some basis or frame. For example, let $x\in\mathbb{C}^n$ be a (perhaps very high dimensional) vector which denotes our signal of interest.  Suppose that for some fixed (known) matrix $D$ whose columns are $n$-dimensional vectors $d_i$, $x$ may be expressed as a linear combination of the columns of $D$, as
\begin{equation}\label{eqn:sparse_repn}
x = \sum_{i} \alpha_i d_i,
\end{equation} 
where the $\alpha_i$ are the coefficients corresponding to the relative weight of the contribution of each of the $d_i$ in the representation. The dictionary $D$ may, for example, consist of all of the columns of an orthonormal matrix (eg., a discrete wavelet or Fourier transform matrix), though other representations may be possible (eg., $D$ may be a frame).  In any case, we define the \emph{support set} $\mathcal{S}$ to be the set of indices corresponding to the nonzero values of $\alpha_i$ in the representation of $x$.  When  $|\mathcal{S}|$ is small relative to the ambient dimension $n$, we say that the signal $x$ is \emph{sparse} in the dictionary $D$, and we call the vector $\alpha$, whose entries are the coefficients $\alpha_i$, the sparse representation of $x$ in the dictionary $D$.

The most general CS observation model prescribes collecting (noisy) linear measurements of $x$ in the form of projections of $x$ onto a set of  $m (<n)$ ``test vectors'' $\phi_i$. Formally, these measurements can be expressed as
\begin{equation}\label{eqn:seq_meas}
y_i = \phi_i^T x + w_i, \ \ i=1,2,\dots,m,
\end{equation}
where $w_i$ denotes the additive measurement uncertainty associated with the $i$th measurement. In ``classic'' CS settings, the measurements are \emph{non-adaptive} in nature, meaning that the $\{\phi_i\}$ are specified independently of $\{y_i\}$ (eg., the test vectors can be specified before any measurements are obtained).  Initial breakthrough results in CS establish that for certain choices of the test vectors, or equivalently the matrix $\Phi$ whose rows are the test vectors, sparse vectors $x$ can be exactly recovered (or accurately approximated) from $m \ll n$ measurements.  For example, if $x$ has no more than $k$ nonzero entries, and the entries of the test vectors/matrix are chosen as iid realizations of zero-mean random variables having sub-Gaussian distributions, then only $m = O(k \log(n/k))$ measurements of the form \eqref{eqn:seq_meas} suffice to exactly recover (if noise free) or accurately estimate (when $w_i\neq 0$) the unknown vector $x$, with high probability \cite{Candes06c, Donoho06}.

Several extensions to the traditional CS paradigm have been investigated recently in the literature.  One such extension  corresponds to exploiting additional \emph{structure} that may be present in the sparse representation of $x$, which can be quantified as follows.  Suppose that $\alpha\in \mathbb{R}^p$, the sparse representation of $x$ in an $n\times p$ orthonormal dictionary $D$, has $k$ nonzero entries.  Then, there are generally $p \choose k$ possible subspaces on which $x$ could be supported, and the space of all $k$-sparse vectors can be understood as a \emph{union of ($k$-dimensional) linear subspaces} \cite{Lu}.  \emph{Structured sparsity} refers to sparse representations that are drawn from a \emph{restricted} union of subspaces (where only a subset of the $p\choose k$ subspaces are allowable).  Recent works exploiting structured sparsity in CS reconstruction include \cite{Huang:09:Structured, Baraniuk:10:ModelBased}. One particular example of structured sparsity, which will be our primary focus here, is \emph{tree-sparsity}.  Let $\mathcal{T}_{p,d}$ denote a balanced rooted connected tree of degree $d$ with $p$ nodes.  Suppose that the components of a sparse representation $\alpha\in\mathbb{R}^p$ can be put into a one-to-one correspondence with the nodes of the tree $\mathcal{T}_{p,d}$.  We say that the vector $\alpha\in\mathbb{R}^p$ is $k$\emph{-tree-sparse} in the tree $\mathcal{T}_{p,d}$ if its nonzero components correspond to a rooted connected subtree of $\mathcal{T}_{p,d}$. This type of tree structure arises, for example, in the wavelet coefficients of many natural images \cite{Crouse}. 

Another extension to the ``classic'' CS observation model is to allow additional flexibility in the \emph{measurement} process in the form of feedback.  Sequential \emph{adaptive sensing} strategies are those for which subsequent test vectors $\{\phi_i\}_{i\geq j}$ may explicitly depend on (or be a function of) past measurements and test vectors $\{\phi_l, y_l\}_{l < j}$.  Adaptive CS procedures have been shown to provide an improved resilience to noise relative to traditional CS -- see, for example, \cite{Ji:08:BCS, Castro:08:Needles, Haupt:09:CDS}, as well as the summary article \cite{Haupt:11:Chapter} and the references therein.  The essential idea of these sequential procedures is to gradually ``steer'' measurements towards the subspace in which signal $x$ resides, in an effort to increase the signal to noise ratio (SNR) of each measurement. 

In this paper we examine a hybrid technique to exploit structured sparsity and adaptivity in the context of noisy compressed sensing. Adaptive sensing techniques that exploit the hierarchical tree-structured dependencies present in wavelet representations of images have been examined in the context of non-Fourier encoding in magnetic resonance imaging \cite{Panych:94:Wavelet}, and more recently in the context of compressed sensing for imaging \cite{SAS2009}. Our first contribution here is to quantify the performance of such procedures when measurements are corrupted by zero-mean additive white Gaussian measurement noise. Our main theoretical results establish sufficient conditions (in terms of the number of measurements required, and the minimum amplitude of the nonzero components) under which the support of tree-sparse vectors may be exactly recovered (with high probability) using these adaptive sensing techniques.  Our results stand in stark contrast with existing results for support recovery for (generally unstructured) sparse vectors, highlighting the significant improvements that can be achieved by the intelligent exploitation of structure throughout the measurement process.   

Further, we demonstrate that tree-based adaptive compressed sensing strategies can be applied with representations learned from a collection of \emph{training data} using recent techniques in hierarchical dictionary learning. This procedure of learning structured sparse representations gives rise to a powerful general-purpose sensing and reconstruction method, which we refer to as \textbf{L}earning \textbf{A}daptive \textbf{Se}nsing \textbf{R}epresentations, or \textbf{LASeR}.  We demonstrate the performance improvements that may be achieved via this approach, relative to other compressed sensing methods.

The remainder of this paper is organized as follows. Section~\ref{sec:sensing} provides a discussion of the top down adaptive compressed sensing procedure motivated by the approaches in \cite{Panych:94:Wavelet, SAS2009}, and contains our main theoretical results which quantify the performance of such approaches in noisy settings.  In Section~\ref{sec:laser} we discuss the LASeR approach for extending this adaptive compressed sensing idea to general compressed sensing applications using recent techniques in dictionary learning.  The performance of the LASeR procedure is evaluated in Section~\ref{sec:evaluation}, and conclusions and directions for future work are discussed in Section~\ref{sec:conclusions}.  Finally, a sketch of the proof of our main result is provided in Section~\ref{sec:proof}.

\section{Adaptive CS for Tree Sparse Signals}\label{sec:sensing}

Our analysis here pertains to a simple adaptive compressed sensing procedure for tree sparse signals, similar to the techniques proposed in \cite{Panych:94:Wavelet, SAS2009}.  As above, let $\alpha\in\mathbb{R}^p$ denote the tree-sparse representation of an unknown signal $x\in\mathbb{R}^n$ in a known $n \times p$ dictionary $D$ having orthonormal columns.  We assume sequential measurements of the form specified in \eqref{eqn:seq_meas} where the additive noises $w_i$ are taken to be iid $\mathcal{N}(0,1)$.  

Rather than projecting onto randomly generated test vectors, here we will obtain measurements of $x$ by projecting onto selectively chosen, scaled versions of columns of the dictionary $D$, as follows.  Without loss of generality suppose that the index $1$ corresponds to the root of the tree $\mathcal{T}_{p,d}$. Begin by initializing a data structure (a stack or queue) with the index $1$, and collect a (noisy) measurement of the coefficient $\alpha_1$ according to \eqref{eqn:seq_meas} by selecting $\phi_1 = \beta d_1$, where $\beta > 0$ is a fixed scaling parameter.  That is, obtain a measurement
\begin{equation}
y = \beta d_1^Tx + w.
\end{equation}
Note that our assumptions on the additive noise imply that $y\sim\mathcal{N}(\beta \alpha_1, 1)$. Now, perform a significance test to determine whether the amplitude of the measured value $y$ exceeds a specified threshold $\tau > 0$. If the measurement is deemed significant (ie, $|y|\geq \tau$), then add the locations of the $d$ children of index $1$ in the tree $\mathcal{T}_{p,d}$ to the stack (or queue). If the measurement is not deemed significant, then obtain the next index from the data structure (if the structure is nonempty) to determine which column of $D$ should comprise the next test vector, and proceed as above. If the data structure is empty, the procedure stops.  Notice that using a stack as the data structure results in \emph{depth-first} traversal of the tree, while using a queue results in \emph{breadth-first} traversal. The aforementioned algorithm is adaptive in the sense that the decision on which locations of $\alpha$ to measure depends on outcomes of the statistical tests corresponding to the previous measurements.

The performance of this procedure is quantified by the following result, which comprises the main theoretical contribution of this work. A sketch of the proof of the theorem is given in Sec.~\ref{sec:proof}.
\begin{theorem}\label{thm:main}\emph{
Let $\alpha$ be $k$-tree-sparse in the tree $\mathcal{T}_{p,d}$ with support set $\mathcal{S}$, and suppose $k<p/d$.  For any $c_1>0$ and $c_2\in(0,1)$, there exists a constant $c_3>0$ such that if
\begin{equation}
\alpha_{\rm min} = \min_{i\in\mathcal{S}} |\alpha_i| \geq \sqrt{c_3 \frac{\log k}{\beta^2}}
\end{equation}
and $\tau = c_2 \beta \alpha_{\rm min}$, the following hold with probability at least $1-k^{-c_1}$: the total number of measurements obtained $m=dk + 1$, and the support estimate $\widehat{S}$ comprised of all the measured locations for which corresponding measured value exceeds $\tau$ in amplitude is equal to $\mathcal{S}$.}
\end{theorem}

A brief discussion is in order here to put the results of this theorem in context.  Note that in practical settings, physical constraints (eg., power or time limitations) effectively impose a limit on the precision of the measurements that may be obtained.  This can be modeled by introducing a global constraint of the form
\begin{equation}
\sum_i \|\phi_i\|_2^2 \leq R,
\end{equation}
on the model \eqref{eqn:seq_meas} in order to limit the ``sensing energy'' that may be expended throughout entire measurement process.  In the context of Thm.~\ref{thm:main}, this corresponds to a constraint of the form $\sum_{i=1}^m \beta^2 \leq R$.  In this case for the choice
\begin{equation}
\label{eq:6}
\beta = \sqrt{\frac{R}{(d+1)k}},
\end{equation}
Thm.~\ref{thm:main} guarantees exact support recovery with high probability from $O(k)$ measurements provided that $\alpha_{\rm min}$ exceeds a constant times $\sqrt{(d+1) (k/R) \log k}$.  To assess the benefits of exploiting structure via adaptive sensing, it is illustrative to compare the result of Thm.~\ref{thm:main} with results obtained in several recent works that examined support recovery for \emph{unstructured} sparse signals under a Gaussian noise model.  The consistent theme identified in these works is that exact support recovery is impossible unless the minimum signal amplitude $\alpha_{\rm min}$ exceeds a constant times $\sqrt{(n/R) \log n}$ for non-adaptive measurement strategies \cite{Donoho:04:HC, Genovese}, or $\sqrt{(n/R) \log k}$ for adaptive sensing strategies \cite{Malloy}.  Clearly, when the signal being acquired is sparse ($k << n$), the procedure analyzed in this work succeeds in recovering much weaker signals.

Our proof of Thm.~\ref{thm:main} can be extended to obtain guarantees on the accuracy of an estimate obtained via a related adaptive sensing procedure.
\begin{corollary}\emph{
There exists a two-stage (support recovery, then estimation) adaptive compressed sensing procedure for $k$-tree sparse signals that produces an estimate from $m=O(k)$ measurements that (with high probability) satisfies
\begin{equation}
\|\hat{\alpha}-\alpha\|_2^2 = O\left(k \left(\frac{k}{R}\right)\right)
\end{equation}
provided $\alpha_{\rm min}$ exceeds a constant times $\sqrt{(k/R) \log k}$.}
\end{corollary}
By comparison, non-adaptive CS estimation techniques that do not assume any structure in the sparse representation can achieve estimation error
\begin{equation}
||\alpha - \hat{\alpha}||_2^2 = O\left(k\left(\frac{n}{R}\right)\log n\right),
\end{equation}
from $m=O(k\log (n/k))$ measurements \cite{Candes:07:Dantzig}.  Exploiting structure in non-adaptive CS, as in \cite{Baraniuk:10:ModelBased}, results in an estimation procedure that achieves error 
\begin{equation}
||\alpha - \hat{\alpha}||_2^2 = O\left(k\left(\frac{n}{R}\right) \right)
\end{equation}
from $m=O(k)$ measurements. Again, we see that the results of the corollary to Thm.~\ref{thm:main} provide a significant improvement over these existing error bounds, especially in the case when $k \ll n$.

\section{Learning Adaptive Sensing Representations}\label{sec:laser}

The approach outlined above can be applied in general settings, by employing techniques from \emph{dictionary learning} \cite{Olshausen:97:Sparse, Aharon:06:KSVD}.  Let $X$ denote an $n \times q$ matrix whose $n$-dimensional columns $x_i$ comprise a collection of training data, and suppose we can find a factorization of $X$ of the form $X \approx DA$, where $D$ is an $n\times p$ dictionary with orthonormal columns, and $A$ is a $p\times q$ matrix whose columns $a_i\in\mathbb{R}^p$ each exhibit tree-sparsity in some tree $\mathcal{T}_{p,d}$.  The task of finding the dictionary $D$ and associated coefficient matrix $A$ with tree-sparse columns can be accomplished by solving an optimization of the form
\begin{equation}\label{eqn:tree_learn}
\{D,A\} = \arg \min_{D \in \mathbb{R}^{n\times q}, \{a_i\}\in\mathbb{R}^{q}} \sum_{i=1}^p\|x_i-Da_i\|_2^2 + \lambda\Omega(a_i),
\end{equation}
subject to the constraint $D^TD = I$.  Here, the regularization term is given by
\begin{equation}
\Omega(a_i) = \sum_{g\in\mathcal{G}} \omega_g \|(a_i)_g\|,
\end{equation}
where $\mathcal{G}$ is the set of $p$ groups, each comprised of a node with all of its descendants in the tree $\mathcal{T}_{p,d}$, the notation $(a_i)_g$ refers to the subvector of $a_i$ restricted to the indices in the set $g\in\mathcal{G}$, the $\omega_g$ are non-negative weights, and the norm can be either the $\ell_2$ or $\ell_{\infty}$ norm. Efficient software packages have been developed (eg., \cite{Jenatton:10:Proximal}) for solving the optimizations of the form \eqref{eqn:tree_learn} via alternating minimization over $D$ and $A$.  Enforcing the additional constraint of orthogonality of the columns of $D$ can be achieved in a straightforward manner.  In the context of the procedure outlined in Sec.~\ref{sec:sensing}, we refer to solving this form of constrained structured dictionary learning task as \textbf{L}earning \textbf{A}daptive \textbf{Se}nsing \textbf{R}epresentations, or \textbf{LASeR}.  The performance of LASeR is evaluated in the next section.

\section{Experimental Results}\label{sec:evaluation}

We performed experiments on the Psychological Image Collection at Stirling \cite{PICS} which contains a set of $72$ man-made and $91$ natural images. The files are in JPG and TIFF format respectively, with each image of size $256 \times 256$ (here, each of the images was rescaled to $128 \times 128$ to reduce computational demands on the dictionary learning procedure). The training data were then each reshaped to a ${16384 \times 1}$ vector and stacked together to form the training matrix $ X \in\mathbb{R}^{16384 \times 163}$. After centering the training data by subtracting the column mean of the training matrix from each of the training vectors, we learned a balanced binary tree structured orthonormal dictionary with $7$ levels (comprising $127$ orthogonal dictionary elements).

\begin{figure*}[thb]
\centering
\begin{tabular}{ccc}
\epsfig{file=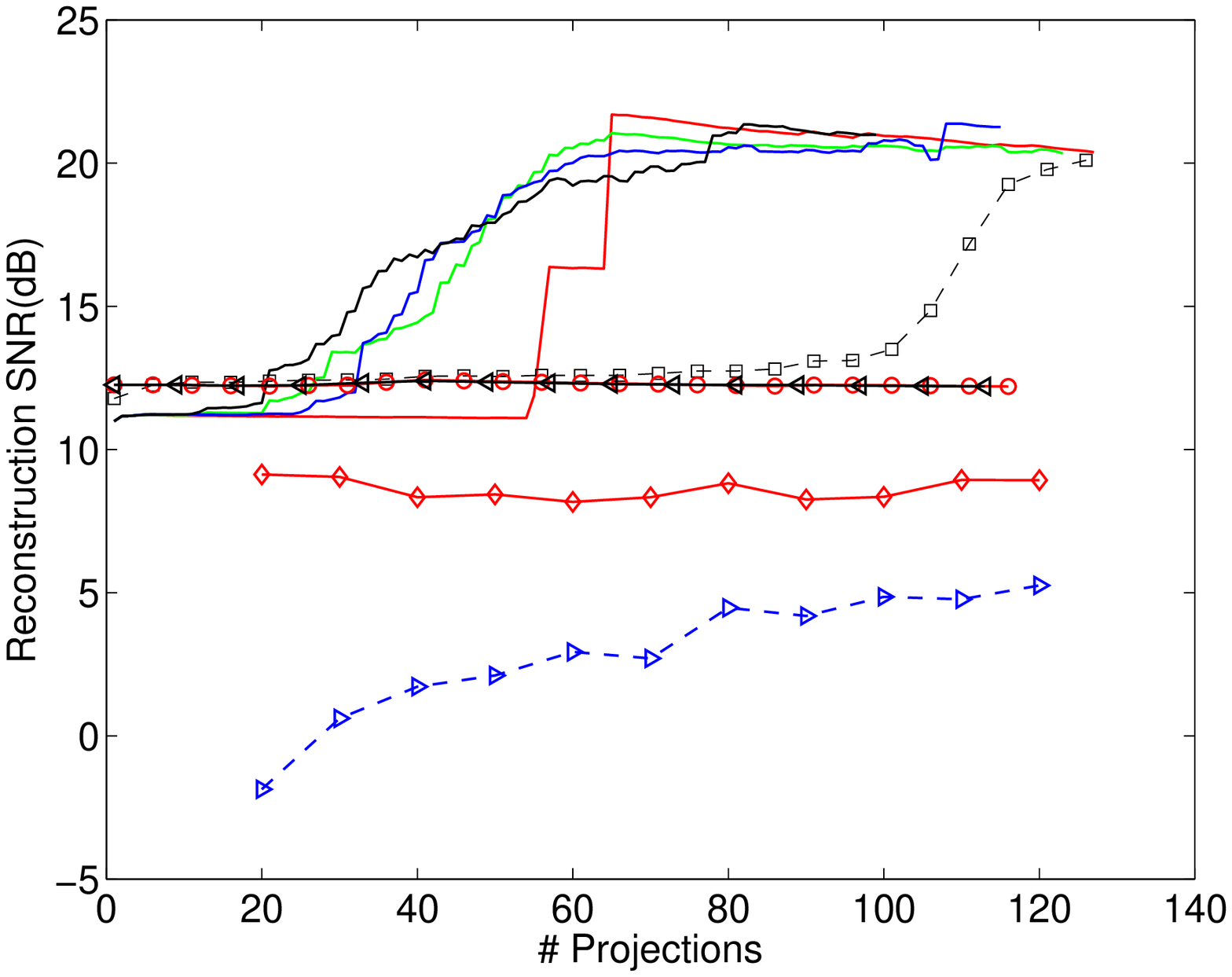,width=0.31\linewidth,clip=} &
\epsfig{file=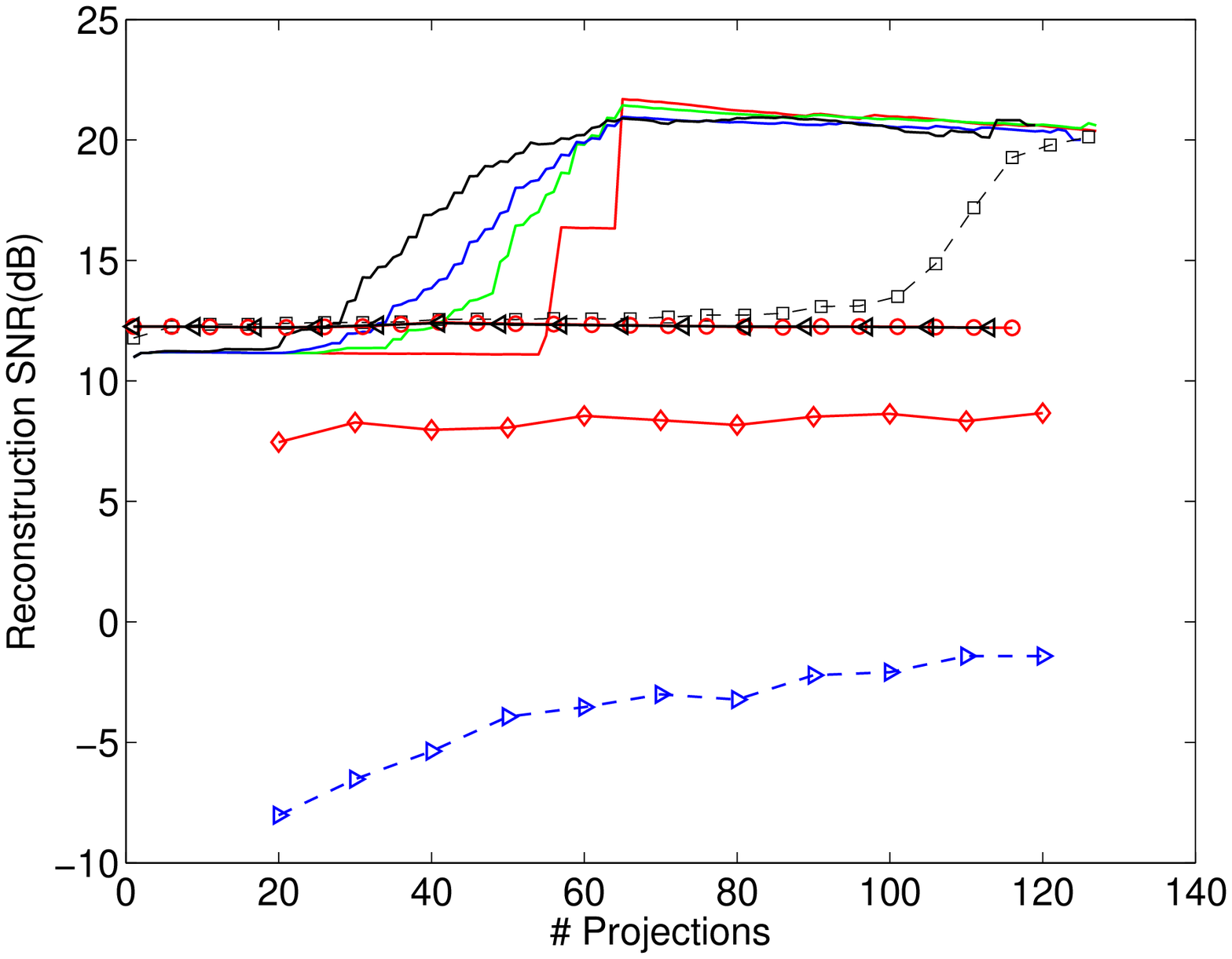,width=0.31\linewidth,clip=} &
\epsfig{file=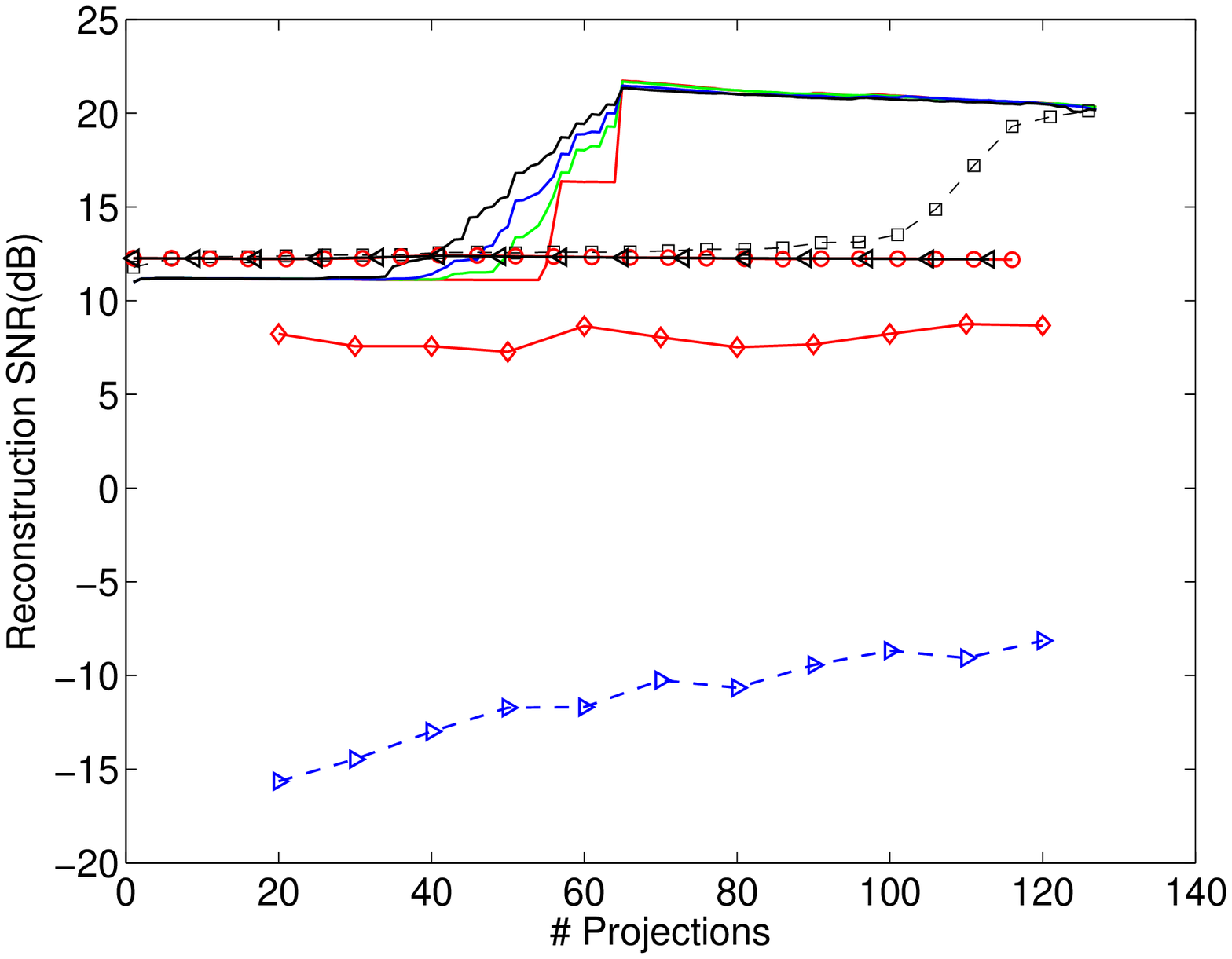,width=0.31\linewidth,clip=}\\
\epsfig{file=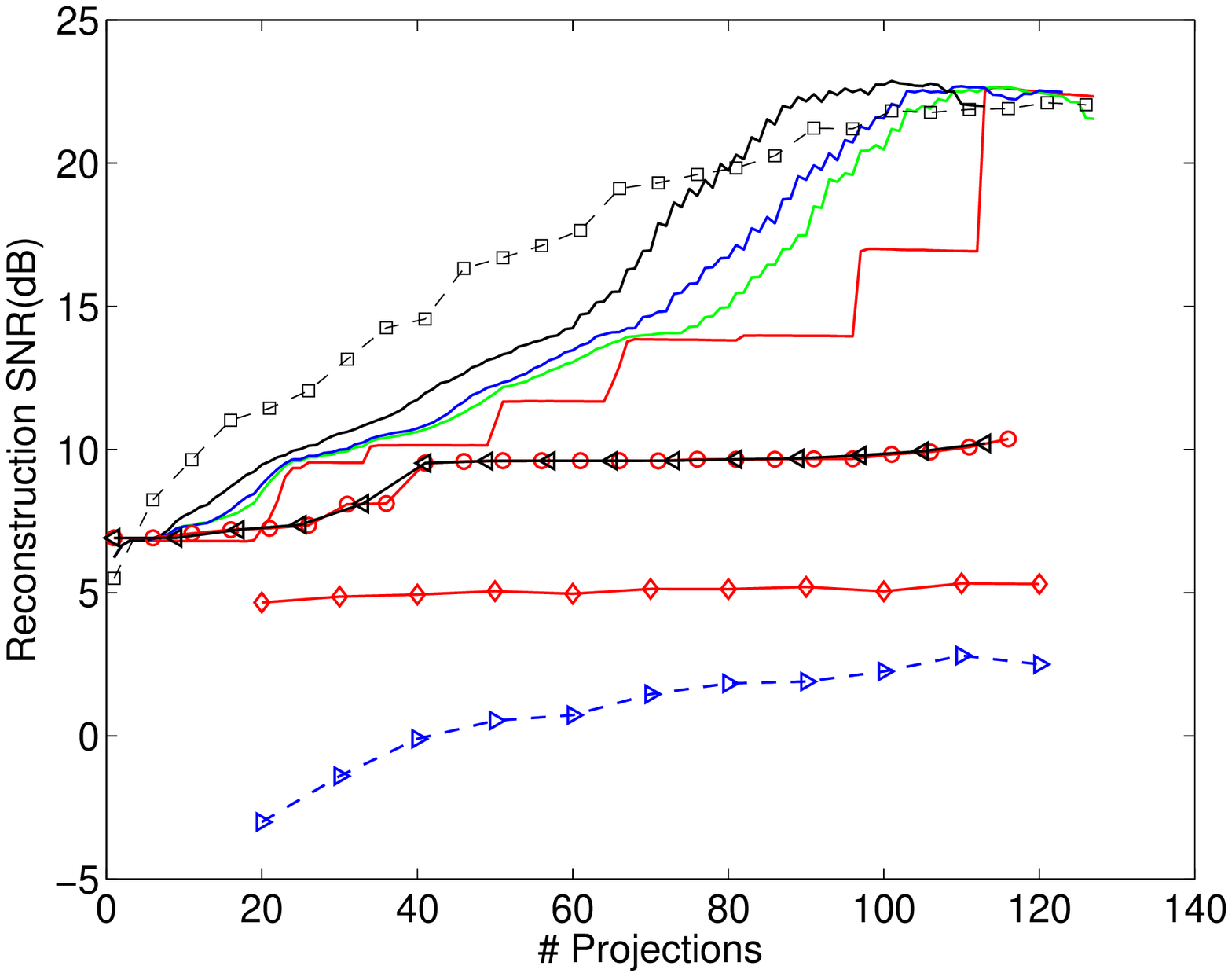,width=0.31\linewidth,clip=} &
\epsfig{file=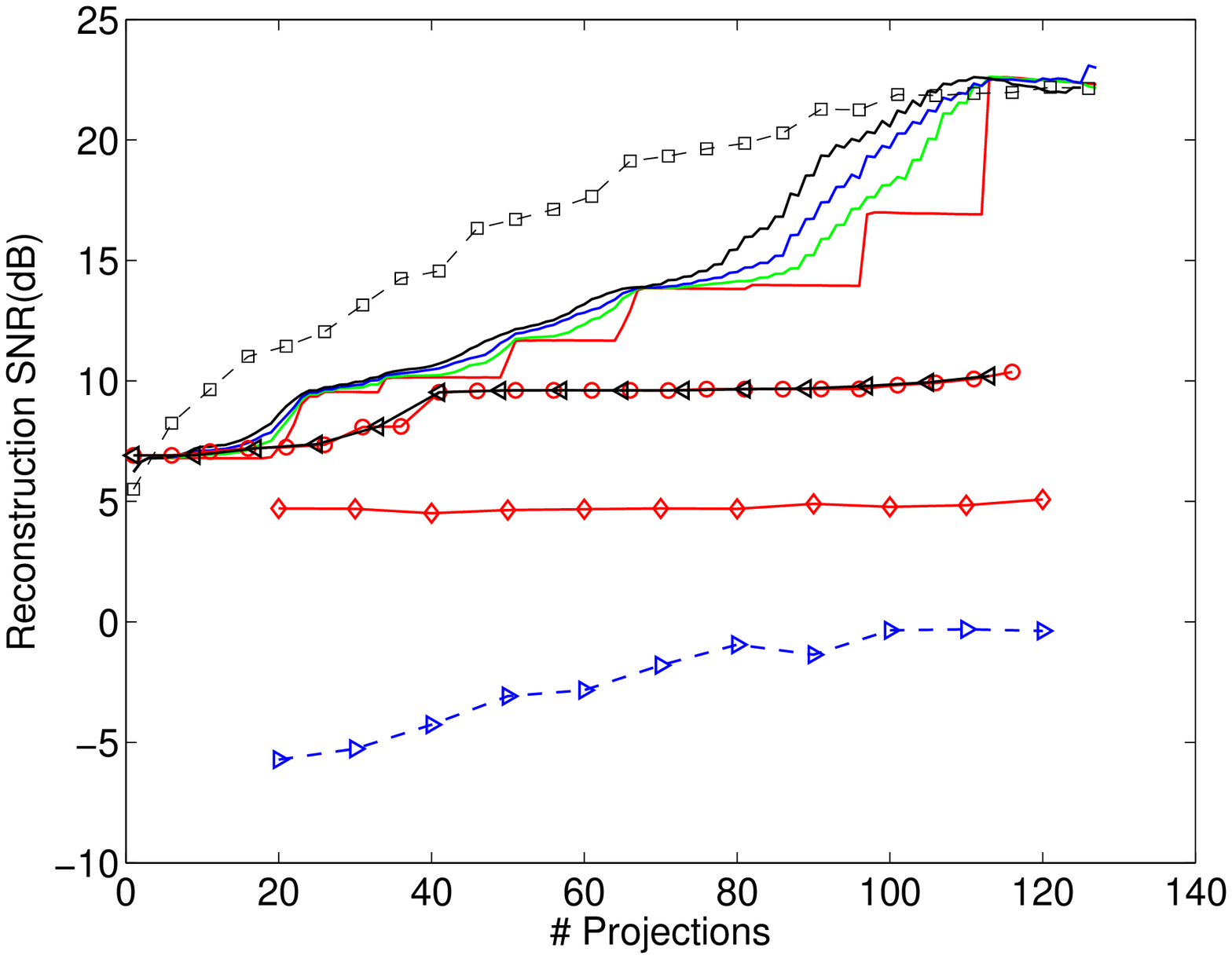,width=0.31\linewidth,clip=} &
\epsfig{file=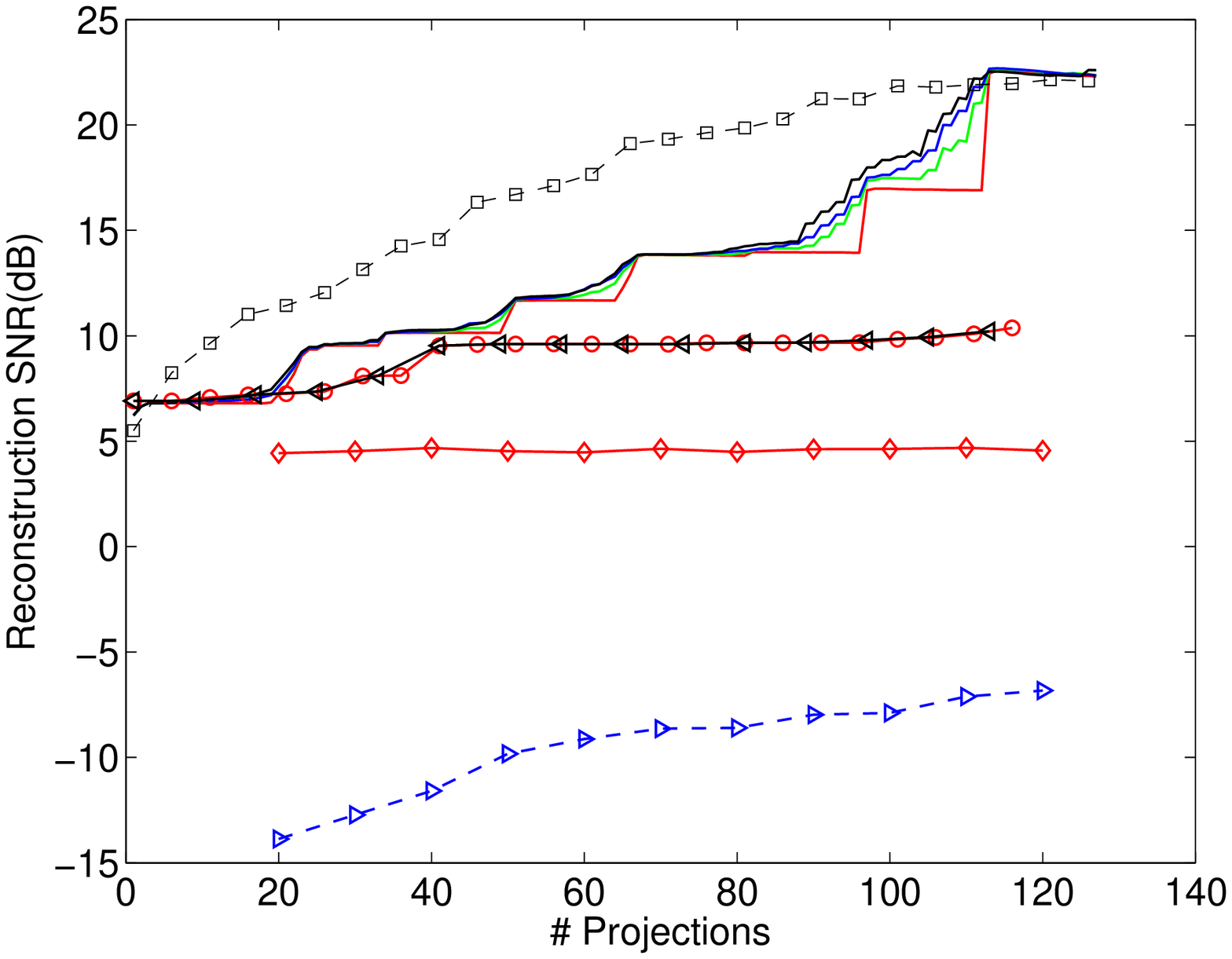,width=0.31\linewidth,clip=}
\end{tabular}
\caption{Reconstruction SNR vs. Number of measurements plots (best viewed in color) with different sensing energy $R$ and fixed noise level $\sigma^2 = 1$ for different schemes (LASeR, PCA, direct wavelet sensing, model-based CS and Lasso). Results in each row corresponds to a different test image. Column $1$: $R = 128\times 128$, Column $2$: $R = (128\times 128)/8$, Column $3$: $R = (128\times 128)/32$. Here, $\medsquare$ is PCA, $\diamond$ is model-based CS, $\triangleright$ is CS-Lasso, $\triangleleft$ and $\circ$ are for direct wavelet sensing with $\tau = 0$ and $\tau = 0.5$ respectively. Colored solid lines are for LASeR with red for $\tau = 0$, green for $\tau = 0.04$, blue for $\tau = 0.06$ and black for $\tau = 0.1$.}
\label{result}
\end{figure*}

The LASeR sensing procedure was then applied with rows of dictionary scaled to meet the total sensing budget $R$ for two test signals (chosen from the original training set). Since, during the dictionary learning process we specify the sparsity level of the signal in the learned dictionary, allocation of sensing energy to each measurement can be done beforehand (specifically $\beta$ is defined as in $(\ref{eq:6})$). We evaluated the performance of the procedure for various values of $\tau$ (the threshold for determining significance of a measured coefficient) in a noisy setting corrupted by zero-mean additive white Gaussian measurement noise.  The reconstruction from the LASeR procedure is obtained as the column mean plus a weighted sum of the atoms of the dictionary used to obtain the projections, where the weights are taken to be the actual observation values obtained by projecting onto the corresponding atom. When assessing the performance of the procedure in noisy settings, we averaged performance over a total of $500$ trials corresponding to different realizations of the random noise.

Reconstruction performance is quantified by the reconstruction signal to noise ratio (SNR), given by 
\begin{equation}
\mbox{SNR}=10\log_{10}\left(\frac{||\rm{x}||^{2}_{2}}{||\hat{\rm{x}} - \rm{x}||^{2}_{2}}\right).
\end{equation}
where $\rm{x}$ and $\rm{\hat{x}}$ are the original test and reconstructed signal respectively.

To provide a performance comparison for LASeR, we also evaluate the reconstruction performance of the direct wavelet sensing algorithm described in \cite{SAS2009}, as well as Principal component analysis (PCA) based reconstuction.  For PCA, the reconstruction is obtained by taking projections of the test signal onto the principal components and adding back the subtracted column mean to the reconstruction. We also compare with ``traditional'' compressed sensing and model-based compressed sensing \cite{Baraniuk:10:ModelBased}, where measurements are obtained by projecting onto random vectors (in this case, vectors whose entries are i.i.d.~zero-mean Gaussian distributed) and reconstruction is obtained via the Lasso and CoSaMP respectively.  In order to make a fair comparison among all of the different strategies, we scale so that the constraint on the total sensing energy is met.

Reconstruction SNR values vs. number of measurements for two of the test images is shown in Fig. \ref{result}.  The results in the top row (for the first test image) show that for a range of threshold values $\tau$ one can get a good reconstruction SNR by taking only $60 - 65$ measurements using LASeR with very limited sensing budget $R$. On the other hand, reconstruction SNR for Lasso and model-based CS degrade as we decrease the sensing energy $R$.  The results in the bottom row (corresponding to the second test image) demonstrate a case where the performance of LASeR is on par with PCA. In this case too, the SNR for Lasso and model-based CS decrease significantly as we decrease $R$. The advantage of LASeR is in the low measurement  (high threshold)  and low sensing budget scenario where we can get a good reconstruction from few measurements. 

\section{Discussion/Conclusion}
\label{sec:conclusions}
In this paper, we presented a novel sensing and reconstruction procedure called LASeR, which uses dictionaries learned from training data, in conjunction with adaptive sensing, to perform compressed sensing. Bounds on minimum feature strength in the presence of measurement noise were explicitly proven for LASeR. Simulations demonstrate that the proposed procedure can provide significant improvements over traditional compressed sensing (based on random projection measurements), as well as other established methods such as PCA.  

Future work in this direction will entail obtaining a complete characterization of the performance of the LASeR procedure for different dictionaries, and for different learned tree structures (we restricted attention here to binary trees, though higher degrees can also be obtained via the same procedure.

\section{Proof of Main Result}\label{sec:proof}

Before proceeding with the proof of the main theorem, we state an intermediate result concerning the number of measurements that are obtained via the procedure described in Sec.~\ref{sec:sensing} when sensing a $k$-tree-sparse vector.  We state the result here as a lemma. The proof is by induction on $k$, and is straightforward, so we omit it here due to space constraints.

\begin{lemma}\emph{
Let $\mathcal{T}_{p,d}$ denote a completed rooted connected tree of degree $d$ with $p$ nodes, and let $\alpha\in\mathbb{R}^p$ be $k$-tree-sparse in $\mathcal{T}_{p,d}$ with $k \leq q/d$.  If the procedure described in Sec.~\ref{sec:sensing} is used to acquire $\alpha$, and the outcome of the statistical test is correct at each step, then the procedure halts when $m=dk + 1$ measurements have been collected.}
\end{lemma}

In other words, suppose that for a $k$-tree sparse $\alpha$ with support $\mathcal{S}$, the outcomes of each of the statistical tests of the procedure described in Sec.~\ref{sec:sensing} are correct.  Then, the set of locations that are measured is of the form $\mathcal{S} \cup \widetilde{\mathcal{S}}^c$, where $\mathcal{S}$ and $\widetilde{\mathcal{S}}^c$ are disjoint, and $|\widetilde{\mathcal{S}}^c| = (d-1)k+1$.

\subsection{Sketch of Proof of Theorem~\ref{thm:main}}

An error can occur in two different ways, corresponding to missing a truly significant signal component (a \emph{miss hit}) and determining a component to be significant when it is not (a \emph{false alarm}).  Let $y_{d_j}$ correspond to the measurement obtained according to the noisy linear model \eqref{eqn:seq_meas} by projecting onto the column $d_j$. A false alarm corresponds to the event $|y_{d_j}| \geq \tau$ for some $j\in\widetilde{\mathcal{S}}^c$. Since in this case, we have $y_{d_j}\sim\mathcal{N}(0,1)$, using a standard Gaussian tail bound for zero mean and unit variance random variables, the probability of false alarm can be upper bounded as
\begin{equation}
\mathrm{Pr}(\mathrm{false~alarm}) \leq e^{-\tau^2/2}.
\end{equation}
Likewise, for $j\in\mathcal{S}$, a miss hit corresponds to $|y_{d_j}| < \tau$.  
Letting $\alpha_{\rm min} = \min_{j\in\mathcal{S}} |\alpha_j|$, we have 
\begin{equation}
\mathrm{Pr}(\mathrm{miss~hit})\leq e^{-(\beta \alpha_{\rm min}-\tau)^2/2}
\end{equation}
for $\tau < \beta\alpha_{\rm min}$. 

Now, the probability of exact support recovery corresponds to the probability of the event that each of the $|\mathcal{S}| = k$ statistical tests corresponding to measurements of nonzero signal components is correct, as are each of the $|\widetilde{\mathcal{S}}^c| = m-k = (d-1)k + 1$ tests corresponding to measurements obtained at locations where the signal has a zero component. Thus, the probability of the failure event can be obtained via the union bound, as
\begin{eqnarray}
\nonumber
\mathrm{Pr}(\mathrm{failure})&\leq& |\widetilde{\mathcal{S}}^c|~ \mathrm{Pr}(\mathrm{false~alarm})+|\mathcal{S}|~ \mathrm{Pr}(\mathrm{miss~hit})\\
&\leq& (m-k)e^{-\tau^2/2}+ke^{-(\beta \alpha_{\rm min}-\tau)^2/2}\label{eqn:bound}
\end{eqnarray}
Let $\tau = a(\beta \alpha_{\rm{min}})$, where $a \in (0,1)$. If, for some $c_1>0$, each of the terms in the bound above is less than $k^{-c_1}/2$, then the overall failure probability is upper bounded by $k^{-c_1}$.  

Consider the first term on the right hand side of \eqref{eqn:bound}, the condition $(m-k)e^{-\tau^2/2} \leq k^{-c_1}/2$ implies that (for $m=dk+1$), 
\begin{equation}\label{eqn:cond1}
\alpha_{\rm min} \geq \sqrt{\frac{2 \log{\left((d-1)k + 1\right)} + 2c_1\log{k} + 2\log{2}}{\beta^2 a^2}}.
\end{equation}
Similarly, the condition $ke^{-(\beta \alpha_{\rm min}-\tau)^2/2} \leq k^{-c_1}/2$ implies
\begin{equation}\label{eqn:cond2}
\alpha_{\rm min} \geq \sqrt{\frac{2(1+c_1)\log{k} + 2\log{2}}{\beta^2(1-a)^2}}.
\end{equation}
There exists a constant $c_3$ (depending on $d$ and $a$) such that when
$\alpha_{\rm min} \geq \sqrt{c_3 \log (k)/\beta^2}$, both \eqref{eqn:cond1} and \eqref{eqn:cond2} are satisfied.

\bibliographystyle{IEEEbib}
\bibliography{bibfile}

\end{document}